\documentclass[11pt]{article}
\usepackage{emnlp2021}
\usepackage{times}
\usepackage{latexsym}

\usepackage[ruled,vlined]{algorithm2e}
\usepackage[T1]{fontenc}
\usepackage[utf8]{inputenc}

\usepackage{amsmath}
\usepackage{cleveref}       
\usepackage{url}            % simple URL typesetting
\usepackage{booktabs}       % professional-quality tables
\usepackage{mathtools}
\usepackage{amssymb}
\usepackage{bm}
\usepackage{bbm}
\usepackage{multirow}
\usepackage{amsfonts}       % blackboard math symbols
\usepackage{nicefrac}       % compact symbols for 1/2, etc.
\usepackage{microtype}
\usepackage{enumitem}
\usepackage{mathtools}
\usepackage{fancybox}

\newcommand{\argmax}{\text{argmax}}% # COLOR
\definecolor{pinegreen}{rgb}{0.0, 0.47, 0.44}
\definecolor{olive}{rgb}{0.5, 0.5, 0.0}
\definecolor{ao}{rgb}{0.0, 0.5, 0.0}
\definecolor{darkpastelgreen}{rgb}{0.01, 0.75, 0.24}
\definecolor{forestgreen}{rgb}{0.13, 0.55, 0.13}
\definecolor{htmlgreen}{rgb}{0.0, 0.5, 0.0}

\crefformat{section}{\S#2#1#3} % see manual of cleveref, section 8.2.1
\crefformat{subsection}{\S#2#1#3}
\crefformat{subsubsection}{\S#2#1#3}

\title{Reward Optimization for Neural Machine Translation with Learned Metrics}

\author{Raphael Shu \textsc{,} Kang Min Yoo and
  Jung-Woo Ha\\ NAVER AI Lab
  \\ 
  \texttt{shu@deeplearn.org}, \texttt{\{kangmin.yoo, jungwoo.ha\}@navercorp.com}}

\date{}

\begin{document}
\maketitle

%%%%%%%%%%%%%%%%%%%%%%%%%%%%%%%%%%%%%%%%
\begin{abstract}
Neural machine translation (NMT) models are conventionally trained with token-level negative log-likelihood (NLL), which does not guarantee that the generated translations will be optimized for a selected sequence-level evaluation metric. 
%In practice, token-level NLL loss is effective at achieving high translation quality when evaluated with BLEU scores. 
Multiple approaches are proposed to train NMT with BLEU as the reward, in order to directly improve the metric. However, it was reported that the gain in BLEU does not translate to real quality improvement, limiting the application in industry. Recently, it became clear to the community that BLEU has a low correlation with human judgment when dealing with state-of-the-art models. %\cite{Mathur2020TangledUI}. 
This leads to the emerging of model-based evaluation metrics. These new metrics are shown to have a much higher human correlation. In this paper, we investigate whether it is beneficial to optimize NMT models with the state-of-the-art model-based metric, BLEURT. %\cite{Sellam2020BLEURTLR}. 
We propose a contrastive-margin loss for fast and stable reward optimization suitable for large NMT models. In experiments, we perform automatic and human evaluations to compare models trained with smoothed BLEU and BLEURT to the baseline models. Results show that the reward optimization with BLEURT is able to increase the metric scores by a large margin, in contrast to limited gain when training with smoothed BLEU. The human evaluation shows that models trained with BLEURT improve adequacy and coverage of translations. Code is available via \url{https://github.com/naver-ai/MetricMT}.
\end{abstract}
%%%%%%%%%%%%%%%%%%%%%%%%%%%%%%%%%%%%%%%%

%%%%%%%%%%%%%%%%%%%%%%%%%%%%%%%%%%%%%%%%
\section{Introduction}
\label{intro}
Although human evaluation of machine translation (MT) systems provides extensive insights such as adequacy, fluency, and comprehensibility \cite{White1994TheAM}, it is costly and time-consuming. In contrast, automatic evaluation provides a quick and simple measure of translation quality, which can be used for comparing multiple MT systems.  Until recently, automatic MT evaluation relies on matching-based metrics such as BLEU \cite{Papineni2002BleuAM}, Meteor \cite{Banerjee2005METEORAA}, and RIBES \cite{Isozaki2010AutomaticEO}. These metrics judge the quality of translations by comparing them to a set of reference translations. BLEU, the most popular metric in MT community, is a document-level statistical measure that evaluates each translation using $n$-gram matching rate.

However, multiple recent studies \cite{CallisonBurch2006ReevaluationTR,Mathur2020TangledUI} show that BLEU and other matching-based metrics have limited performances when judging among the state-of-the-art systems. In particular, \citet{Mathur2020TangledUI} demonstrates that when considering only a few top systems in WMT'19 metrics task \cite{ma-etal-2019-results}, the human correlation of almost all automatic metrics drop to zero or negative territory. BLEU is further shown to have the worst performance among several competitors. Consequently, the authors call for stopping the use of BLEU and its variants. Indeed, \citet{CallisonBurch2006ReevaluationTR} has already warned the risk of BLEU, and provided insights on when BLEU can go wrong. First, BLEU is weak at permutation. Given a translation, we can permute the $n$-grams and still get similar BLEU scores. In practice, we may observe BLEU assigns a high score to a translation with unusual word orders. Second, the non-matching phrases with no support in the reference are the gray zone of BLEU. For non-matching phrases, BLEU is incapable of detecting the existence of a critical mistake that alters the meaning completely.

Recently, multiple automatic metrics based on trained neural models have emerged. For example, ESIM \cite{Chen2017EnhancedLF} is a model based on bi-LSTM and average pooling, which can be tuned for predicting human judgment scores of translation quality. Yisi-1 \cite{Lo2019YiSiA} considers the distributional lexical similarity. BERTscore \cite{Zhang2020BERTScoreET} evaluates the aggregated pairwise cosine similarity of token embeddings given a hypothesis and a reference. All these models either utilize BERT token embeddings or build based on BERT \cite{Devlin2019BERTPO}. Studies have shown that the model-based metrics indeed have higher human correlation. \citet{Mathur2020TangledUI} reports that the model-based metrics outperform BLEU in almost all situations.

Given the significant success of recent development on model-based metrics, we are motivated to examine whether it is beneficial to train or tune NMT models to directly optimize these new metrics. There are two folds of benefits. First, the model-based metrics provide a smoother training signal that covers the gray zone of non-matching phrases, where BLEU is less effective. Second and most importantly, when optimizing a learned metric~\footnote{Note that not all model-based metrics are trained to predict human scores (e.g., BERTscore).} that is trained to match human judgments, it may adjust NMT parameters to reflect human preferences on different aspects of translation quality.

In this paper, we report results of optimizing NMT models with the state-of-the-art learned metric, BLEURT \cite{Sellam2020BLEURTLR}. BLEURT utilizes synthetic data for BERT pretraining, where the BERT is asked to predict multiple criteria including BLEU, BERTscore, and likelihoods of back-translation. Then, it is tuned with human scores provided by WMT metrics shared tasks (2017-2019). Specifically, we want to investigate whether optimizing a learned metric yields meaningful change to the translations, or it just learns to hack the BERT model with meaningless sequences. If the changes are meaningful, then we want to know what aspects of the translations are improved.

Given that both the BERT-based BLEURT and Transformer-based NMT models are large models with high GPU memory consumption, to facilitate fast and stable reward optimization, we propose to use a pairwise ranking loss that differentiates the best and worst sequence in a candidate space. Unlike the conventional risk loss \cite{Shen2016MinimumRT}, it does not require evaluating the model scores of all candidates, therefore has a smaller memory footprint.

We evaluate optimized NMT models on four language pairs: German-English, Romanian-English, Russian-English, and Japanese-English. Experiments show that when using smoothed BLEU as the reward, we gained limited improvement on BLEU scores, which is in line with past researches \cite{Edunov2018ClassicalSP}. However, when optimizing BLEURT, we observe significant gains in BLEURT scores. In all datasets, the BLEURT scores are increased by more than 10\% on a relative scale. The human evaluation shows that reward optimization with BLEURT tends to improve the adequacy and coverage of translations. Finally, we present a qualitative analysis on the divergence between BLEURT and BLEU. The contributions of this paper can be summarized as:
\begin{enumerate}[itemsep=0ex,partopsep=1ex,parsep=1ex]
\item We propose a constrastive-margin loss that enables fast and stable reward optimization with large NMT and metric models
\item We found that the reward optimization with BLEURT is able to boost the score significantly. However, it hurts the BLEU scores. 
\item The human evaluation shows that the resultant translations, even with lower BLEU, tend to have better adequacy and coverage.
\end{enumerate}

\section{Reward Optimization for Sequence-level Training}

Given a source-target sequence pair $(X, Y^*)$ from the training set, let $\mathcal{S}(X, \theta, N)$ be a set of $N$ candidate sequences generated by model $p_{\theta}(Y|X)$, which is reachable by a certain decoding strategy (e.g. beam search). For neural machine translation, the goal of reward optimization is to solve the following bi-level optimization (BLO) problem:
\begin{align}
   & \:\:\:\:\:\:\:\:\:\:\:\:\:\:\: \max_{\theta} R(\hat Y), \label{eq:ul_problem} \\
    \text{s.t.} \:\:\:\:\:  & \hat Y =  \mathop{\argmax}\limits_{Y \in \mathcal{S}(X, \theta, N)} \log p_\theta(Y|X). \label{eq:ll_problem}
\end{align}
where $R(\cdot)$ is the reward of a sequence. Bi-level optimization \cite{Dempe2002FoundationsOB,Colson2007AnOO} is a framework that formulates such a constrained optimization problem. The solution of the upper-level (UL) problem is constrained to be the best response (i.e., argmax solution) of the lower-level (LL) problem. In the case of neural machine translation, the UL problem is to maximize a specific metric. The LL problem constrains the solution to be the candidate in the reachable set that has the highest NMT model score. Here, we assume that each candidate has a unique model score, so that the argmax operator will not produce a set. This is referred to as the lower-level singleton assumption \cite{Liu2020AGF}, the BLO problem will be much more complicated without such an assumption. In this paper, we define $R(\hat Y)$ as the {\it best-response reward}. The BLO problem in this paper can be simply described as best-response reward maximization.

Recently, BLO gains popularity in tasks such as hyper-parameter optimization \cite{Maity2019ABA}, architecture search \cite{Souquet2020HyperFDAAB,Hou2020SinglelevelOF} and meta-learning \cite{Bertinetto2019MetalearningWD}, where the problem is solved by constructing the explicit gradient $\nabla_\theta R$ \cite{Grazzi2020ConvergencePO,Finn2017ModelAgnosticMF,Okuno2018HyperparameterLV}. These solutions require at least first-order differentiability for the LL problem. However, for machine translation, computing the explicit gradient is extremely difficult given the fact that $R(\cdot)$ can be non-differentiable and $Y$ is a discrete sequence.

Multiple reward optimization objectives including policy gradient and risk minimization have been explored for NMT \cite{Ranzato2016SequenceLT,Shen2016MinimumRT}. In fact, existing approaches can be seen as solutions derived from approximated problems of BLO. We can classify existing approaches into two categories according to the approximation they use.

\paragraph{Expected Reward Maximization} The first category of solutions approximates the best-response reward using the expected reward. We can relax the candidate set constraint so that the LL solution can take any sequence. Then, the original BLO problem can be approximated by maximizing the expected reward:
\begin{align}
J(\theta) = \mathbb{E}_{Y \sim p_\theta(Y|X)} \Big [ R(Y) \Big ]. \label{eq:rl_approx}
\end{align}
This gives the exact form of the policy gradient objective. Here, Eq.~\eqref{eq:rl_approx} describes an one-step markov decision process (MDP) that treats the whole sequence $Y$ as an action. The gradient of this objective can be computed explicitly by applying the log-derivative trick. In the case of autoregressive models, we often model the decoding process as a multi-step MDP \cite{Ranzato2016SequenceLT} with the following objective:
\begin{align}
    J(\theta) = \sum_{t=1}^{T} \mathbb{E}_{y_t \sim p_\theta(y_t|y_{<t}, X)} \Big[ R(y_{<t}, y_t) \Big].
\end{align}
Here, $R(y_{<t}, y_t)$ denotes a partial reward of the currently generated sequence, which requires careful design in order to reflect both the long-term reward and immediate feedback.

We may also choose to retain the candidate set constraint, this leads to the risk loss:
\begin{align}
     \mathcal{L}(\theta) &= \mathbb{E}_{Y \sim \tilde p_\theta(Y|X)} \Big [ -R(Y) \Big ], \nonumber \\
     \tilde p_\theta(Y|X) &= \frac{p_\theta(Y|X)}{\sum_{Y^\prime \in \mathcal{S}(X, \theta, N)} p_\theta(Y^\prime|X)}. \label{eq:risk}
\end{align}
In contrast to the policy gradient objective, the risk objective minimizes the expected cost of a finite number of candidates, each cost is weighted by the normalized model probability $\tilde p_{\theta}(Y|X)$. In literature, \citet{Shen2016MinimumRT} samples the candidates from the NMT model while \citet{Edunov2018ClassicalSP} uses the candidates found by beam search.  Here, we assume the cost is $- R(Y)$, which can also be defined otherly (e.g., 1 - BLEU). As there is a finite number of candidates, the quantity can be computed exactly by $\sum_{Y \in \mathcal{S}} - R(Y) \tilde p_{\theta}(Y|X)$, which is fully differentiable.

\paragraph{Ranking Optimization}
Assume the candidates in $\mathcal{S}(X, \theta, N)$ are sorted by their model scores in descending order. Rather than maximizing the expected reward, the second category of solutions focuses on putting the candidates in the correct order according to their rewards. This is a valid strategy as we can assume that a small change to the parameter $\theta$ will not remove or add new sequences to the candidate set. The only way for the hypothetical gradient $\nabla_\theta R$ to improve the best-response reward is to rerank the candidates so that a candidate with a higher reward will be placed at the top. Therefore, the ranking problem is a proxy problem of the original BLO problem.

Various methods are explored in this category. Here, we denote $Y^*_R = \argmax_{Y \in \mathcal{S}} R(Y)$ as the candidate with the highest reward. The simplest approach is to improve the model score of $Y^*_R$ with:
\begin{align}
    \mathcal{L}(\theta) = - \log p_\theta(Y^*_R|X). \label{eq:reward_distill}
\end{align}
% We refer to this objective as {\it reward distillation}.
When the candidate set has only one sequence ($N = 1$), the objective resorts to self distillation. However, the model may optimize the loss by simply lowering the softmax temperature. To avoid this problem, we can substract Eq.~\eqref{eq:reward_distill} with a baseline and form a margin loss. \citet{Tsochantaridis2005LargeMM,Edunov2018ClassicalSP} described a {\it multi-margin loss} by using the model scores of other candidates as baselines:
\begin{align}
    \mathcal{L}(\theta) &=\smashoperator{\sum_{Y \in \mathcal{S} \setminus \{Y^*_R\} }} \max \Big( 0, m - s_\theta(Y^*_R|X) + s_\theta(Y|X)\Big), \nonumber \\
    m &= \alpha \Big( R(Y^*_R) - R(Y) \Big).
\end{align}
Here, $s_{\theta}(\cdot)$ denotes the logits. The margin $m$ is computed with the reward difference, scaled by a hyperparameter $\alpha$. The authors also described a {\it max-margin loss} that uses the best competitor as baseline:
\begin{align}
    \mathcal{L}(\theta) &= \max \Big( 0, m - s_\theta(Y^*_R|X) + s_\theta(\hat Y|X)\Big),\nonumber \\
    \hat Y &= \mathop{\argmax}\limits_{Y \in \mathcal{S} \setminus \{Y^*_R\}} p_\theta(Y|X),  \nonumber\\
    m &= \alpha \Big( R(Y^*_R) - R(\hat Y) \Big). \label{eq:maxmargin}
\end{align}

%%%%%%%%%%%%%%%%%%%%%%%%%%%%%%%%%%%%%%%%
\section{Contrastive-Margin Loss for Reward Optimization}
As the policy gradient method optimizes the expected reward without the candidate set constraint, it may diverge from best-response reward maximization. Also, because it is difficult to construct the reward in the multi-step MDP setting, the risk loss is favored in recent papers. \citet{Shen2016MinimumRT} and \citet{Edunov2018ClassicalSP} found that the risk loss performs the best among various structured prediction objectives. However, the risk loss in Eq.~\eqref{eq:risk} requires computing the model scores for all sequences in the candidate set, which has the size of at least 10 in practice. This significantly slows down the training especially in our setting where both the reward function and NMT model are large neural networks.  The max-margin loss in Eq.~\eqref{eq:maxmargin} is the only objective that has a small memory footprint as it only computes the model scores of two candidates.

However, when optimizing the max-margin loss with BLEURT, we found the training to be unstable. When $Y^*_R$ has a low rank, the gradient of the max-margin loss still penalizes the argmax solution $\hat Y$ even when $\hat Y$ actually has a relatively high reward. As a result, a third candidate with a worse reward may be placed at the top.
% - This can be monitored by ...

In this paper, we propose to use a more conservative variation of the max-margin loss, which we refer to as {\it contrastive-margin loss}. Denote $Y^{\sim}_R$ as the candidate with the worst reward in $\mathcal{S}(X, \theta, N)$. The constrastive-margin loss differentiates between the best and worst candidate in the set with:
\begin{align}
    \mathcal{L}(\theta) &= \max \Big( 0, m - s_\theta(Y^*_R|X) + s_\theta(Y^{\sim}_R|X)\Big),\nonumber \\
    m &= \alpha \Big( R(Y^*_R) - R(Y^{\sim}_R) \Big). \label{eq:conmargin}
\end{align}
As penalizing $Y^\sim_R$ is a much safer choice compared to penalizing $\hat Y$, the training will be more stable. Although the proposed loss is not optimizing the best-response reward $R(\hat Y)$ explicitly, we observe that both the best-response reward and the average reward are improved over time.

The size of the candidate set $N$ will impact the behavior of Eq.~\eqref{eq:conmargin}. When $N=1$, the loss produces zero gradient as $Y^*_R = Y^\sim_R$. When $N$ increases, the training becomes more stable as the $Y^*_R$ and $Y^\sim_R$ will be less likely to change when the candidate set is large enough.

%%%%%%%%%%%%%%%%%%%%%%%%%%%%%%%%%%%%%%%%
\section{Related Work}
Indeed, fine-tuning machine translation systems with the target metric is a standard practice in statistical machine translation \cite{Och2003MinimumER,Cer2008RegularizationAS}. For neural machine translation, existing researches have attempted to improve rule-based rewards (e.g., BLEU) with reinforcement learning \cite{Ranzato2016SequenceLT,Wu2018ASO}, risk minimization \cite{Shen2016MinimumRT,Edunov2018ClassicalSP}, and other methods \cite{Wiseman2016SequencetoSequenceLA,Welleck2020MLEguidedPS}. \citet{Wieting2019BeyondBT} and \citet{Schmidt2020BERTAA} proposed metrics based on the similarity between sentence embeddings, and optimize the text generation models with the similarity scores.

In this paper, we focus on optimizing NMT models with learned metrics (e.g., BLEURT), which are trained with human judgment data because human perceives the translation quality in multiple criteria (e.g., adequacy, fluency and coverage) with different weights. If a single metric cannot accurately match human preference, inevitably, we have to mix different rewards and tune the weights with human annotators. In contrast, a learned metric might provide a balanced training signal that reflects human preference.

Recent model-based metrics are all built using BERT, which can guide the models to produce semantically sound translations. Different from our approach, \citet{Zhu2020IncorporatingBI} and \citet{Chen2019DistillingTK} attempt to directly incorporate BERT into NMT models. However, reward training with learned metrics can benefit more than just semantics, as shown in the qualitative analysis of this paper.

%%%%%%%%%%%%%%%%%%%%%%%%%%%%%%%%%%%%%%%%
\section{Experiments}

\subsection{Settings}

\begin{table}[!t]
\small
\centering
\begin{tabular}{ccccc} \toprule
 & De-En & Ro-En & Ru-En & Ja-En \\
\midrule
domain & Web & News & Web & Science \\
\#train & 4.5M & 608K & 1M & 2.2M \\
\#test & 3003 & 1999 & 2000  & 1812 \\
avg. words & 28.5 & 26.4 & 24.6 & 25.7 \\
\bottomrule
\end{tabular}
\caption{Corpus statistics including domain, number of training and testing pairs and average number of words}
\label{table:dataset}
\end{table}

\paragraph{Dataset}
We conduct experiments on four to-English datasets: WMT14 De-En \cite{Bojar2014FindingsOT}, WMT16 Ro-En \cite{Bojar2016FindingsOT}, WMT19 Ru-En \cite{Barrault2019FindingsOT} and ASPEC Ja-En \cite{Nakazawa2016ASPECAS}. Table~\ref{table:dataset} reports the corpus statistics. All datasets are preprocessed using Moses tokenizer except the Japanese corpus, which is tokenized with Kytea \cite{Neubig2011PointwisePF}. Bype-pair encoding \cite{Sennrich2016NeuralMT} with a code size of 32K is applied for all training data. For De-En and Ro-En, we follow the default practice in \verb|fairseq| to bind all embeddings.  For WMT19 Ru-En task, we only use Yandex Corpus~\footnote{\url{https://translate.yandex.ru/corpus}} as training data. For ASPEC Ja-En corpus, we perform data filtering to remove abnormal pairs, where the number of source and target words differ by more than 50\%.

\paragraph{Model}
We use Base Transformer \cite{vaswani17attention} for De-En, Ru-En, and Ja-En tasks. For Ro-En task, we reduce the number of hidden units to 256 and feed-forward layer units to 1024. Other hyperparameters follow the default setting of \verb|fairseq| \cite{Ott2018ScalingNM}. We train the baseline models for 100K iterations and average the last 10 checkpoints. For Ro-En, however, we observe over-fitting before the training ends. Therefore, we average 10 checkpoints with the last checkpoint having the lowest validation loss. Most experiments were performed on NAVER Smart Machine Learning (NSML)~\cite{kim2018nsml}.

\begin{figure}[t]
\centering
  \includegraphics[width=\linewidth]{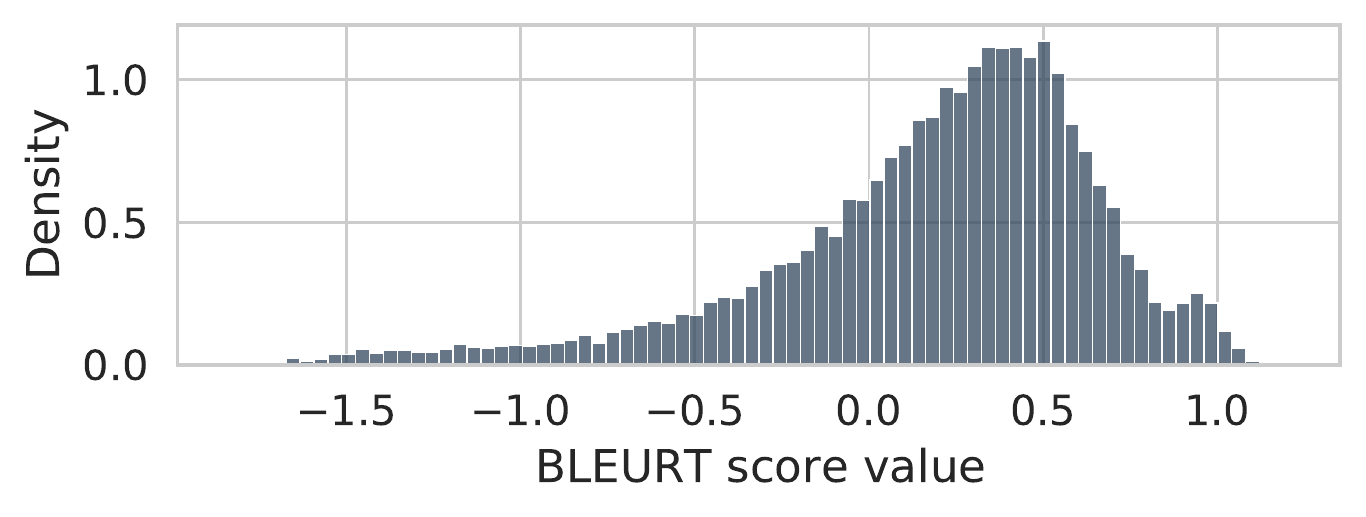}
\caption{Histogram of BLEURT scores produced by baseline Transformers on mixed language pairs}
\label{fig:bleurt_dist}
\end{figure}

\paragraph{Reward}
In our experiments, we test with two rewards: Smoothed BLEU (SLBEU) \cite{SBLEU} and BLEURT \cite{Sellam2020BLEURTLR}. For computing the BLEURT score, we use pretrained BLEURT-Base model~\footnote{\url{https://github.com/google-research/bleurt}} that has 12 layers with 768 hidden units, allowing a maximum of 128 tokens. As BLEURT predicts the quality score with a dense layer, the value of scores does not fall into a certain range. Fig.~\ref{fig:bleurt_dist} shows the histogram of BLEURT scores produced using mixed unique hypothesis-reference pairs sampled from all datasets. The translations are generated using baseline Transformers. The BLEURT scores have a mean of 0.18 and a standard deviation of 0.48.

\begin{figure}[t]
\centering
  \includegraphics[width=\linewidth]{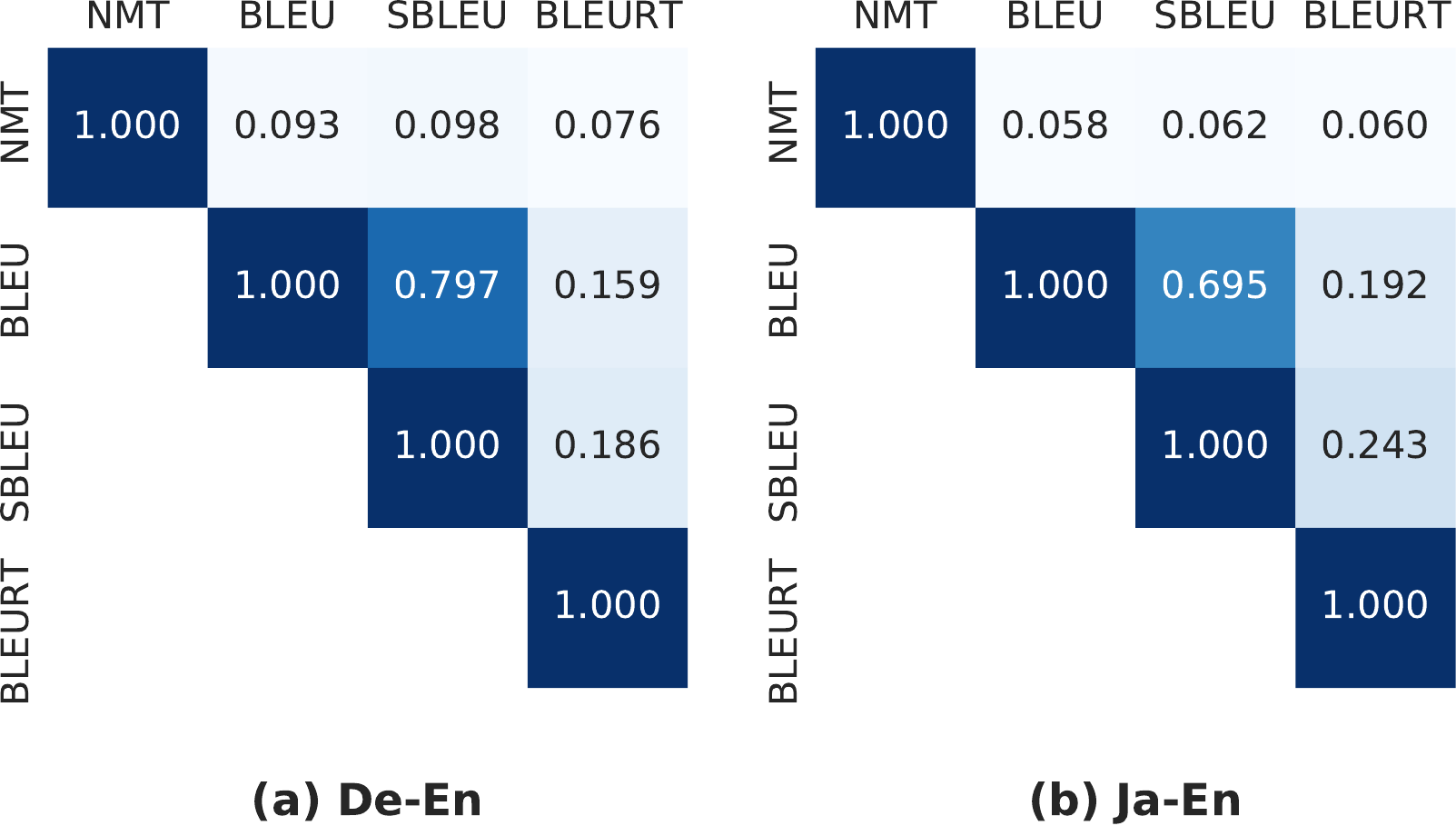}
\caption{Correlation (Kendall's $\tau$) between different automatic metrics. The coefficients are computed over $4$-best candidates then averaged. NMT indicates the model score of baseline Transformers.}
\label{fig:agreement}
\end{figure}

\begin{figure*}[t]
\centering
  \includegraphics[width=\linewidth]{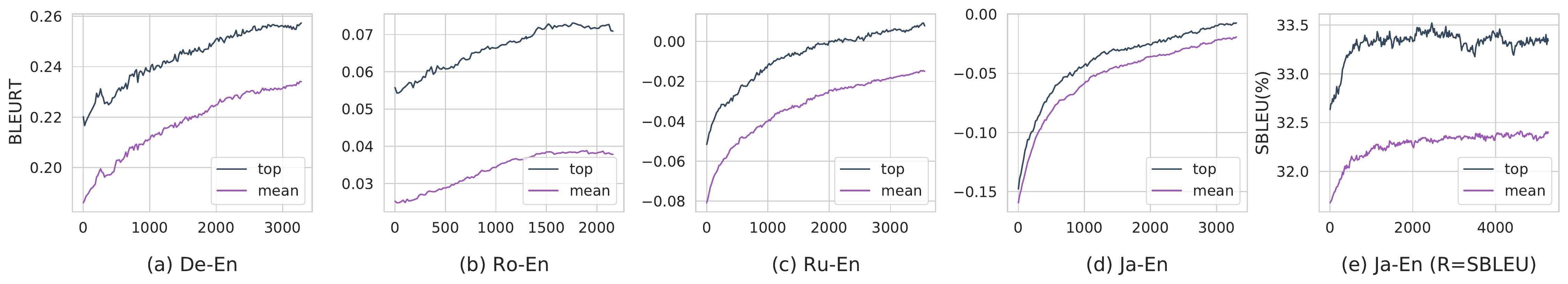}
\caption{Validation reward (BLEURT) over iterations (X axis) when fine-tuning baseline Transformers using the contrastive-margin loss (Eq.~\eqref{eq:conmargin}). The upper line in each plot shows the reward value of the top candidate. The lower line shows the mean reward value of top-10 candidates produced by beam search. The last plot shows the training graph with SBLEU as the reward.}
\label{fig:reward_opt}
\end{figure*}

\paragraph{Training and Hyperparameters}
We start with the baseline model parameters and use the contrastive-margin loss to finetune the parameters. We use SGD with a learning rate of $0.0001$. In each training iteration, we perform beam search to produce $N$ candidates. As the beam size in the inference time is 4, we set $N=10$ so that the training can be stable. The margin hyperparameter $m$ in Eq.~\eqref{eq:conmargin} is chosen based on validation reward. We found that $m=0.3$ generally works for SBLEU and $m=0.1$ generally works for BLEURT in all datasets. As both the BLEURT model and the Transformer have high GPU memory consumption, we set a batch size of 100 sentences for BLEURT and 1024 tokens for the NMT model. The training is distributed on 16 Nvidia V100 GPUs, which typically converges within $1 \sim 2$ epochs.
In our experiments, it takes around 80 minutes for running 1000 iterations. We check the validation reward for every 20 iterations and save the checkpoint with the highest best-response reward.

\subsection{Agreement between Different Metrics}

In Fig.~\ref{fig:agreement}, we show the correlation between different metrics on German-English (a) and Japanese-English (b) tasks. We measure Kendal's $\tau$ of metric pairs of $4$-best candidate translations sampled from baseline Transformers. We report the coefficients averaged over 10K random samples. In the figure, we also show the correlation of baseline NMT model scores (i.e., length-normalized log probability) and other automatic metrics.

The correlation matrix reveals that when evaluating the top few candidates produced by beam search, the baseline NMT model scores have extremely low correlation with both evaluation metrics we consider. Therefore, if optimizing a certain metric is beneficial, then there exists huge room for improvement. Results also show that BLEURT has a weak correlation with BLEU/SBLEU.

\subsection{Effects of Reward Optimization}

Fig.~\ref{fig:reward_opt} demonstrates the effects on validation rewards when fine-tuning baseline NMT models with the proposed contrastive-margin loss (Eq.~\eqref{eq:conmargin}). In each plot, we show the best-response reward (top) and the average reward of top-10 candidates (mean) found by beam search. When optimizing BLEURT (Fig.~\ref{fig:reward_opt} (a-d)), we observe the reward training to be stable and the validation reward is improved significantly over time. For the low-resource dataset Ro-En, we observe overfitting after training for 2000 iterations.

When training with SBLEU as the reward, however, we observe the validation reward to be less stable. In the case of Ja-En (Fig.~\ref{fig:reward_opt} (e)), the SLBEU of the top candidate reaches the maximum value after around 2000 iterations. For SBLEU, two sequences will end up with the same score if they contain only a few different $n$-grams that do not match the reference. Unfortunately, due to the low diversity of beam search, this is often the case. As a result, the number of effective training samples will be much less than the batch size.  The contradiction of training behavior indicates that the smoothness of the reward has a large impact on the stability of training.

\subsection{Automatic Evaluation Results}

In Table~\ref{table:main}, we report the automatic evaluation scores for NMT models optimized with different rewards and compare them with baseline models. All translations are generated with beam search with a beam size of 4. We report tokenized BLEU and BLEURT-Base scores for all language pairs.

We observe that when optimizing the BLEU scores, the gain is limited. This result is in line with past researches \cite{Edunov2018ClassicalSP}. However, when optimizing BLEURT as the reward, the metric score is significantly improved. In the case of Ja-En, the BLEURT score is relatively increased by $31\%$ when optimizing for it. In contrast, the BLEU score is only improved by $0.3\%$ when trained with SBLEU. For De-En, Ro-En and Ru-En tasks, BLEURT training results in relative improvement of $18\%$ , $11\%$ and $80\%$ respectively.

The second observation is that the BLEU score starts to hurt when BLEURT is improved by a large margin, which happens for De-En, Ru-En, and Ja-En tasks. This indicates that BLEURT has a drastically different preference on translations comparing to BLEU. This phenomenon is particularly significant on Ja-En task, where the BLEU score is reduced by as much as 8 points. To understand how the translations change to enhance BLEURT and even harm BLEU , we perform further qualitative analysis, which is summarized in Section \ref{sec:QA}.

\begin{table*}[t]
\small
\centering
\begin{tabular}{r|cc|cc|cc|cc} \toprule
 \multirow{2}{*}{Model} & \multicolumn{2}{c|}{De-En} & \multicolumn{2}{c|}{Ro-En} & \multicolumn{2}{c|}{Ru-En} &\multicolumn{2}{c}{Ja-En} \\
 & BLEU(\%)  & BLEURT & BLEU(\%)  & BLEURT & BLEU(\%)  & BLEURT & BLEU(\%)  & BLEURT \\
\midrule
Baseline & 30.21 & 0.167 & 34.11 & 0.069 & 34.20 & 0.086 & 28.81 & 0.154\\
R - SBLEU & 30.98 & 0.164 & 34.47 & 0.071 & 35.11 & 0.094 & 28.92 & 0.174\\
R - BLEURT & 28.94 & 0.195 & 34.17 & 0.077 & 33.07 & 0.155 & 20.75 & 0.202\\
\bottomrule
\end{tabular}
\caption{Automatic evaluation scores of the baseline Transformer and models optimized with different rewards}
\label{table:main}
\end{table*}

%%%%%%%%%%%%%%%%%%%%%%%%%%%%%%%%%%%%%%%%
\subsection{Human Evaluation}

We perform human evaluations to examine how the reward training affects the real translation quality. We conduct pairwise judgment with Amazon Mechanical Turk, asking the workers to judge the quality of translations from random model pairs. For each dataset, we submit 50 evaluation tasks for each model pair, each task will be assigned to three different annotators. During our evaluations, the workers are blind to the models. We also add ``reference-baseline'' pair as a control group to verify the validity of collected data, resulting in 600 jobs for each dataset. We filter the workers with an approval rate threshold of 98\% and ensure the workers having at least 1000 approval hits. For pairwise human evaluation, we perform experiments on three major datasets: De-En, Ru-En, and Ja-En. The questionnaire can be found in Appendix.

Following \citet{White1994TheAM}, we ask the annotators to judge the adequacy (i.e. relevancy of meaning) and fluency (i.e. readability) of translations. Additionally, we also ask annotators to judge the coverage of translations. A translation has lower coverage if it has missing meaning. Due to the difficulty of finding workers capable of speaking the source language, we only ask workers to judge by comparing with reference translations. It is optional to check the meaning of the source sentence. On average, workers are spending 57 seconds for completing a job.

The results are summarized in Table~\ref{table:human}. For each criterion, we first count the majority votes for each task. For instance, the result will be a win for ``A-B'' pair on adequacy if two or more workers judge that A's result has better adequacy
than B's result. In the case of all three workers disagree with each other, we discard the sample. The rightmost column of the table shows the average token-level edit distance between the results of the two models. In the last section, we show the combined results across different datasets. 

The token-level edit distance shows that fine-tuning with BLEURT reward changes the translations more aggressively than the SBLEU reward. Based on the combined results, we observe that the BLEURT model tends to outperform both the baseline and the SBLEU model on adequacy and coverage. No significant improvement in fluency is observed. This may indicate the NMT models trained with NLL loss already have good fluency. However, when comparing the models for each dataset in detail, we can see that the only consistently outperforming criterion for the BLEURT model is coverage. The results on adequacy and fluency are mixed depending on the dataset. 

To conclude, human evaluation gives a clear sign that the BLEURT training improves the coverage. Although the improvement on adequacy is uneven, it proves that the translation quality can be improved even with decreased BLEU scores. 

\begin{table*}[t]
\small
\centering
\begin{tabular}{r|r|ccc|ccc|ccc|c} \toprule
\multirow{2}{*}{Dataset} & \multirow{2}{*}{Evaluation Pair} & \multicolumn{3}{c|}{Adequacy} & \multicolumn{3}{c|}{Fluency} & \multicolumn{3}{c|}{Coverage} & \multirow{2}{*}{Edit}  \\
 &  & win & tie & loss & win & tie & loss & win & tie & loss & \\
 \midrule
De-En&bleurt-baseline&{\bf0.21}&0.60&0.19&0.17&0.63&{\bf0.20}&{\bf0.23}&0.65&0.12&4.7 \\
&sbleu-baseline&{\bf0.29}&0.63&0.07&{\bf0.23}&0.70&0.07&{\bf0.23}&0.68&0.09&2.7 \\
&bleurt-sbleu&{\bf0.24}&0.59&0.17&{\bf0.24}&0.55&0.21&0.20&0.55&{\bf0.25}&4.9 \\
% &reference-baseline&{\bf0.60}&0.40&0.00&{\bf0.41}&0.59&0.00&{\bf0.57}&0.43&0.00&12.1 \\
\midrule
Ru-En&bleurt-baseline&{\bf0.26}&0.56&0.19&0.22&0.56&0.22&{\bf0.25}&0.57&0.18&4.3 \\
&sbleu-baseline&{\bf0.31}&0.48&0.21&0.23&0.51&{\bf0.26}&0.26&0.49&0.26&2.1 \\
&bleurt-sbleu&0.24&0.51&0.24&{\bf0.22}&0.59&0.20&{\bf0.27}&0.59&0.15&4.3 \\
% &reference-baseline&{\bf0.57}&0.40&0.02&{\bf0.58}&0.40&0.02&{\bf0.64}&0.36&0.00&11.1 \\
\midrule
Ja-En&bleurt-baseline&0.26&0.45&{\bf0.29}&{\bf0.24}&0.54&0.22&{\bf0.22}&0.58&0.20&12.3 \\
&sbleu-baseline&0.18&0.61&{\bf0.21}&0.18&0.64&0.18&{\bf0.16}&0.71&0.13&5.6 \\
&bleurt-sbleu&{\bf0.33}&0.38&0.29&0.23&0.50&{\bf0.27}&{\bf0.29}&0.45&0.26&11.4 \\
% &reference-baseline&{\bf0.62}&0.38&0.00&{\bf0.55}&0.45&0.00&{\bf0.64}&0.36&0.00&15.5 \\
\midrule
Combined&bleurt-baseline&{\bf0.24}&0.54&0.22&0.21&0.58&0.21&{\bf0.23}&0.60&0.17&7.1 \\
&sbleu-baseline&{\bf0.26}&0.57&0.16&{\bf0.21}&0.62&0.17&{\bf0.22}&0.63&0.16&3.4 \\
&bleurt-sbleu&{\bf0.27}&0.49&0.23&0.23&0.55&0.23&{\bf0.25}&0.53&0.22&6.8 \\
\bottomrule
\end{tabular}
\caption{Pairwise human evaluation results for De-En, Ru-En and Ja-En language pairs. The annotators are asked to judge the adequacy, fluency and coverage of two translations from a random pair of models. We collect three data points for each sample. For each criterion, we show the percentage of majority votes. Last column shows the average token-level edit distance between results. The bottom section shows combined results across datasets.}
\label{table:human}
\end{table*}

%%%%%%%%%%%%%%%%%%%%%%%%%%%%%%%%%%%%%%%%
\subsection{The divergence between BLEU and BLEURT}
\label{sec:QA}

In automatic evaluation, we observed a significant divergence between BLEU and BLEURT. The reward training with BLEURT hurts BLEU in all three major datasets. To understand the factors causing such divergence, we perform qualitative analysis on sampled translations of the reward training model and compare them with the baseline results. We randomly sample 10 translation results for each dataset. Here, We only focus on the results where BLEUT scores strongly disagree with smoothed BLEU. In particular, we only pick translations that increase BLEURT by at least 0.3 and decrease smoothed BLEU by at least 3\%, compared to baseline results. We carefully check the translations to see whether the quality is improved. In the case of improving, we annotate the detailed category of improvement (e.g., semantics, order, coverage, and readability). The result is summarized as below:
\begin{itemize}[itemsep=0ex,partopsep=1ex,parsep=1ex]
    \item \lbrack25/40\rbrack \:\: Improved quality (better semantics 16/40, better order 6/40, better coverage 3/40)
    \item \lbrack10/40\rbrack \:\:  Similar quality
    \item \lbrack\:\:5/40\rbrack \:\:  Degraded quality
\end{itemize}

We found when BLEURT strongly disagree with smoothed BLEU, the majority has improved quality. In most of the cases, we observe the BLEURT-trained model picks words with a semantic meaning closer to the reference. One of such examples is shown in Table \ref{table:example}. Other than semantics, we also found examples showing BLEURT favors the translations with better word order or higher coverage.

\paragraph{Potential Risk} While examining the translations, we also found unexpected results where BLEURT is significantly higher/lower though the translation quality is almost identical with the baseline. Table \ref{table:example} (Ex. 2) demonstrates a constructed example, where the only difference between the two hypotheses is the last name. Although two translations may have a similar quality, the BLEURT scores differ significantly. As BLEURT takes the semantics into consideration, we found that it occasionally assigns unexpected scores when rare words are presented in the translations.

\begin{table}[t]
\small
\centering
\begin{tabular}{p{0.12\linewidth} | p{0.78\linewidth}} \toprule
{\bf Ex 1} & {\bf Improved semantics (Ru-En \#1440)} \\
REF & After the film screening and discussion about the \underline{recent} scientific achievements , ... \\
BASE & After watching the film and discussing \underline{modern} scientific breakthroughs , ... \\
OURS & After watching the film and discussing \underline{the latest} scientific breakthroughs , ... \\
\midrule
{\bf Ex 2} & {\bf Risk relevant to rare words} \\
REF & Billy Donovan will be {sixth} coach of Zach \underline{LaVine} . \\
HYP1 & Donovan will be {fifth} coach of Zach \underline{Curry} . (BLEURT=0.164)\\
HYP2 & Donovan will be {fifth} coach of Zach \underline{Bryant} .  (BLEURT=0.008)\\
\bottomrule
\end{tabular}
\caption{Translation examples that demonstrate improved semantics (Ex. 1) and BLEURT assigning unexpected scores when rare words are presented (Ex. 2). The source sentences are omitted.}
\label{table:example}
\end{table}

\section{Conclusion}

In this paper, we optimize NMT models with the state-of-the-art learned metric, BLEURT, and examine the effects on translation quality with both automatic and human evaluation. To facilitate fast and stable reward training, we propose contrastive-margin loss, which requires computing model scores for only two sequences, in contrast to $N$ sequences in the risk loss. Through experiments, we found optimizing BLEURT results in significant gains in the score. However, in three over four language pairs, the improvement of BLEURT is coupled with degraded BLEU scores. By performing the pairwise human evaluation, we found the NMT models optimized with BLEURT tend to have better adequacy and coverage comparing to baseline and models tuned with smoothed BLEU. The results show that the reward optimization is not simply hacking the metric model with unrealistic translations, but yield meaningful improvement.

Multiple recent studies \cite{Wu2016GooglesNM,Choshen2020OnTW} have indicated that BLEU as a training signal does not contribute to the quality improvement of translations. Based on the results of this paper, we argue reward optimization is still beneficial if we choose the right metric that correctly reflects the human preference of translations. Future work on systematic reward ensembling has the potential to bring consistent improvement to machine translation models.

\bibliography{emnlp2021}

\begin{thebibliography}{47}
\expandafter\ifx\csname natexlab\endcsname\relax\def\natexlab#1{#1}\fi

\bibitem[{Banerjee and Lavie(2005)}]{Banerjee2005METEORAA}
S.~Banerjee and A.~Lavie. 2005.
\newblock Meteor: An automatic metric for mt evaluation with improved
  correlation with human judgments.
\newblock In \emph{IEEvaluation@ACL}.

\bibitem[{Barrault et~al.(2019)Barrault, Bojar, Costa-juss{\`a}, Federmann,
  Fishel, Graham, Haddow, Huck, Koehn, Malmasi, Monz, M{\"u}ller, Pal, Post,
  and Zampieri}]{Barrault2019FindingsOT}
Lo{\"i}c Barrault, Ondrej Bojar, M.~Costa-juss{\`a}, C.~Federmann, M.~Fishel,
  Yvette Graham, B.~Haddow, M.~Huck, Philipp Koehn, S.~Malmasi, Christof Monz,
  Mathias M{\"u}ller, Santanu Pal, Matt Post, and Marcos Zampieri. 2019.
\newblock Findings of the 2019 conference on machine translation (wmt19).
\newblock In \emph{WMT}.

\bibitem[{Bertinetto et~al.(2019)Bertinetto, Henriques, Torr, and
  Vedaldi}]{Bertinetto2019MetalearningWD}
Luca Bertinetto, Jo{\~a}o~F. Henriques, P.~Torr, and A.~Vedaldi. 2019.
\newblock Meta-learning with differentiable closed-form solvers.
\newblock \emph{ArXiv}, abs/1805.08136.

\bibitem[{Bojar et~al.(2014)Bojar, Buck, Federmann, Haddow, Koehn, Leveling,
  Monz, Pecina, Post, Saint-Amand, Soricut, Specia, and
  Tamchyna}]{Bojar2014FindingsOT}
Ondrej Bojar, C.~Buck, C.~Federmann, B.~Haddow, Philipp Koehn, Johannes
  Leveling, Christof Monz, Pavel Pecina, Matt Post, Herve Saint-Amand, Radu
  Soricut, Lucia Specia, and A.~Tamchyna. 2014.
\newblock Findings of the 2014 workshop on statistical machine translation.
\newblock In \emph{WMT@ACL}.

\bibitem[{Bojar et~al.(2016)Bojar, Chatterjee, Federmann, Graham, Haddow, Huck,
  Jimeno-Yepes, Koehn, Logacheva, Monz, Negri, N{\'e}v{\'e}ol, Neves, Popel,
  Post, Rubino, Scarton, Specia, Turchi, Verspoor, and
  Zampieri}]{Bojar2016FindingsOT}
Ondrej Bojar, R.~Chatterjee, C.~Federmann, Yvette Graham, B.~Haddow, M.~Huck,
  Antonio Jimeno-Yepes, Philipp Koehn, V.~Logacheva, Christof Monz, M.~Negri,
  Aur{\'e}lie N{\'e}v{\'e}ol, Mariana Neves, M.~Popel, Matt Post, Rapha{\"e}l
  Rubino, Carolina Scarton, Lucia Specia, M.~Turchi, Karin~M. Verspoor, and
  Marcos Zampieri. 2016.
\newblock Findings of the 2016 conference on machine translation.
\newblock In \emph{WMT}.

\bibitem[{Callison-Burch et~al.(2006)Callison-Burch, Osborne, and
  Koehn}]{CallisonBurch2006ReevaluationTR}
Chris Callison-Burch, M.~Osborne, and Philipp Koehn. 2006.
\newblock Re-evaluation the role of bleu in machine translation research.
\newblock In \emph{EACL}.

\bibitem[{Cer et~al.(2008)Cer, Jurafsky, and Manning}]{Cer2008RegularizationAS}
Daniel~Matthew Cer, Dan Jurafsky, and Christopher~D. Manning. 2008.
\newblock Regularization and search for minimum error rate training.
\newblock In \emph{WMT@ACL}.

\bibitem[{Chen et~al.(2017)Chen, Zhu, Ling, Wei, Jiang, and
  Inkpen}]{Chen2017EnhancedLF}
Qian Chen, Xiao-Dan Zhu, Zhenhua Ling, Si~Wei, Hui Jiang, and D.~Inkpen. 2017.
\newblock Enhanced lstm for natural language inference.
\newblock In \emph{ACL}.

\bibitem[{Chen et~al.(2019)Chen, Gan, Cheng, Liu, and jing
  Liu}]{Chen2019DistillingTK}
Yen-Chun Chen, Zhe Gan, Yu~Cheng, J.~Liu, and Jing jing Liu. 2019.
\newblock Distilling the knowledge of bert for text generation.
\newblock \emph{ArXiv}, abs/1911.03829.

\bibitem[{Choshen et~al.(2020)Choshen, Fox, Aizenbud, and
  Abend}]{Choshen2020OnTW}
Leshem Choshen, Lior Fox, Zohar Aizenbud, and Omri Abend. 2020.
\newblock On the weaknesses of reinforcement learning for neural machine
  translation.
\newblock \emph{ArXiv}, abs/1907.01752.

\bibitem[{Colson et~al.(2007)Colson, Marcotte, and Savard}]{Colson2007AnOO}
B.~Colson, P.~Marcotte, and G.~Savard. 2007.
\newblock An overview of bilevel optimization.
\newblock \emph{Annals of Operations Research}, 153:235--256.

\bibitem[{Dempe(2002)}]{Dempe2002FoundationsOB}
S.~Dempe. 2002.
\newblock Foundations of bilevel programming.

\bibitem[{Devlin et~al.(2019)Devlin, Chang, Lee, and
  Toutanova}]{Devlin2019BERTPO}
J.~Devlin, Ming-Wei Chang, Kenton Lee, and Kristina Toutanova. 2019.
\newblock Bert: Pre-training of deep bidirectional transformers for language
  understanding.
\newblock In \emph{NAACL-HLT}.

\bibitem[{Edunov et~al.(2018)Edunov, Ott, Auli, Grangier, and
  Ranzato}]{Edunov2018ClassicalSP}
Sergey Edunov, Myle Ott, M.~Auli, David Grangier, and Marc'Aurelio Ranzato.
  2018.
\newblock Classical structured prediction losses for sequence to sequence
  learning.
\newblock In \emph{NAACL-HLT}.

\bibitem[{Finn et~al.(2017)Finn, Abbeel, and Levine}]{Finn2017ModelAgnosticMF}
Chelsea Finn, P.~Abbeel, and Sergey Levine. 2017.
\newblock Model-agnostic meta-learning for fast adaptation of deep networks.
\newblock In \emph{ICML}.

\bibitem[{Grazzi et~al.(2020)Grazzi, Pontil, and
  Salzo}]{Grazzi2020ConvergencePO}
Riccardo Grazzi, M.~Pontil, and Saverio Salzo. 2020.
\newblock Convergence properties of stochastic hypergradients.
\newblock \emph{ArXiv}, abs/2011.07122.

\bibitem[{Hou and Jin(2020)}]{Hou2020SinglelevelOF}
Pengfei Hou and Ying Jin. 2020.
\newblock Single-level optimization for differential architecture search.
\newblock \emph{ArXiv}, abs/2012.11337.

\bibitem[{Isozaki et~al.(2010)Isozaki, Hirao, Duh, Sudoh, and
  Tsukada}]{Isozaki2010AutomaticEO}
Hideki Isozaki, T.~Hirao, Kevin Duh, Katsuhito Sudoh, and Hajime Tsukada. 2010.
\newblock Automatic evaluation of translation quality for distant language
  pairs.
\newblock In \emph{EMNLP}.

\bibitem[{Kim et~al.(2018)Kim, Kim, Seo, Kim, Park, Park, Jo, Kim, Yang, Kim
  et~al.}]{kim2018nsml}
Hanjoo Kim, Minkyu Kim, Dongjoo Seo, Jinwoong Kim, Heungseok Park, Soeun Park,
  Hyunwoo Jo, KyungHyun Kim, Youngil Yang, Youngkwan Kim, et~al. 2018.
\newblock Nsml: Meet the mlaas platform with a real-world case study.
\newblock \emph{arXiv preprint arXiv:1810.09957}.

\bibitem[{Lin and Och(2004)}]{SBLEU}
Chin-Yew Lin and F.~Och. 2004.
\newblock Orange: a method for evaluating automatic evaluation metrics for
  machine translation.
\newblock In \emph{COLING}.

\bibitem[{Liu et~al.(2020)Liu, Mu, Yuan, Zeng, and Zhang}]{Liu2020AGF}
R.~Liu, Pan Mu, Xiaoming Yuan, Shangzhi Zeng, and J.~Zhang. 2020.
\newblock A generic first-order algorithmic framework for bi-level programming
  beyond lower-level singleton.
\newblock \emph{ArXiv}, abs/2006.04045.

\bibitem[{kiu Lo(2019)}]{Lo2019YiSiA}
Chi kiu Lo. 2019.
\newblock Yisi - a unified semantic mt quality evaluation and estimation metric
  for languages with different levels of available resources.
\newblock In \emph{WMT}.

\bibitem[{Ma et~al.(2019)Ma, Wei, Bojar, and Graham}]{ma-etal-2019-results}
Qingsong Ma, Johnny Wei, Ond{\v{r}}ej Bojar, and Yvette Graham. 2019.
\newblock \href {https://doi.org/10.18653/v1/W19-5302} {Results of the {WMT}19
  metrics shared task: Segment-level and strong {MT} systems pose big
  challenges}.
\newblock In \emph{Proceedings of the Fourth Conference on Machine Translation
  (Volume 2: Shared Task Papers, Day 1)}, pages 62--90, Florence, Italy.
  Association for Computational Linguistics.

\bibitem[{Maity et~al.(2019)Maity, Maity, and Ghosh}]{Maity2019ABA}
Krishanu Maity, Satyabrata Maity, and Nimisha Ghosh. 2019.
\newblock A bi-level approach for hyper-parameter tuning of an evolutionary
  extreme learning machine.
\newblock \emph{2019 International Conference on Applied Machine Learning
  (ICAML)}, pages 124--129.

\bibitem[{Mathur et~al.(2020)Mathur, Baldwin, and Cohn}]{Mathur2020TangledUI}
Nitika Mathur, Tim Baldwin, and Trevor Cohn. 2020.
\newblock Tangled up in bleu: Reevaluating the evaluation of automatic machine
  translation evaluation metrics.
\newblock In \emph{ACL}.

\bibitem[{Nakazawa et~al.(2016)Nakazawa, Yaguchi, Uchimoto, Utiyama, Sumita,
  Kurohashi, and Isahara}]{Nakazawa2016ASPECAS}
Toshiaki Nakazawa, Manabu Yaguchi, Kiyotaka Uchimoto, M.~Utiyama, E.~Sumita,
  S.~Kurohashi, and H.~Isahara. 2016.
\newblock Aspec: Asian scientific paper excerpt corpus.
\newblock In \emph{LREC}.

\bibitem[{Neubig et~al.(2011)Neubig, Nakata, and Mori}]{Neubig2011PointwisePF}
Graham Neubig, Yosuke Nakata, and Shinsuke Mori. 2011.
\newblock Pointwise prediction for robust, adaptable japanese morphological
  analysis.
\newblock In \emph{ACL}.

\bibitem[{Och(2003)}]{Och2003MinimumER}
F.~Och. 2003.
\newblock Minimum error rate training in statistical machine translation.
\newblock In \emph{ACL}.

\bibitem[{Okuno et~al.(2018)Okuno, Takeda, and
  Kawana}]{Okuno2018HyperparameterLV}
Takayuki Okuno, Akiko Takeda, and Akihiro Kawana. 2018.
\newblock Hyperparameter learning via bilevel nonsmooth optimization.
\newblock \emph{arXiv: Optimization and Control}.

\bibitem[{Ott et~al.(2018)Ott, Edunov, Grangier, and Auli}]{Ott2018ScalingNM}
Myle Ott, Sergey Edunov, David Grangier, and M.~Auli. 2018.
\newblock Scaling neural machine translation.
\newblock In \emph{WMT}.

\bibitem[{Papineni et~al.(2002)Papineni, Roukos, Ward, and
  Zhu}]{Papineni2002BleuAM}
Kishore Papineni, S.~Roukos, T.~Ward, and Wei-Jing Zhu. 2002.
\newblock Bleu: a method for automatic evaluation of machine translation.
\newblock In \emph{ACL}.

\bibitem[{Ranzato et~al.(2016)Ranzato, Chopra, Auli, and
  Zaremba}]{Ranzato2016SequenceLT}
Marc'Aurelio Ranzato, S.~Chopra, M.~Auli, and W.~Zaremba. 2016.
\newblock Sequence level training with recurrent neural networks.
\newblock \emph{CoRR}, abs/1511.06732.

\bibitem[{Schmidt and Hofmann(2020)}]{Schmidt2020BERTAA}
Florian Schmidt and T.~Hofmann. 2020.
\newblock Bert as a teacher: Contextual embeddings for sequence-level reward.
\newblock \emph{ArXiv}, abs/2003.02738.

\bibitem[{Sellam et~al.(2020)Sellam, Das, and Parikh}]{Sellam2020BLEURTLR}
Thibault Sellam, Dipanjan Das, and Ankur~P. Parikh. 2020.
\newblock Bleurt: Learning robust metrics for text generation.
\newblock In \emph{ACL}.

\bibitem[{Sennrich et~al.(2016)Sennrich, Haddow, and
  Birch}]{Sennrich2016NeuralMT}
Rico Sennrich, B.~Haddow, and Alexandra Birch. 2016.
\newblock Neural machine translation of rare words with subword units.
\newblock \emph{ArXiv}, abs/1508.07909.

\bibitem[{Shen et~al.(2016)Shen, Cheng, He, He, Wu, Sun, and
  Liu}]{Shen2016MinimumRT}
Shiqi Shen, Yong Cheng, Zhongjun He, W.~He, Hua Wu, M.~Sun, and Yang Liu. 2016.
\newblock Minimum risk training for neural machine translation.
\newblock \emph{ArXiv}, abs/1512.02433.

\bibitem[{Souquet et~al.(2020)Souquet, Shvai, Llanza, and
  Nakib}]{Souquet2020HyperFDAAB}
Leo Souquet, Nadiya Shvai, Arcadi Llanza, and A.~Nakib. 2020.
\newblock Hyperfda: a bi-level optimization approach to neural architecture
  search and hyperparameters' optimization via fractal decomposition-based
  algorithm.
\newblock \emph{Proceedings of the 2020 Genetic and Evolutionary Computation
  Conference Companion}.

\bibitem[{Tsochantaridis et~al.(2005)Tsochantaridis, Joachims, Hofmann, and
  Altun}]{Tsochantaridis2005LargeMM}
Ioannis Tsochantaridis, T.~Joachims, Thomas Hofmann, and Y.~Altun. 2005.
\newblock Large margin methods for structured and interdependent output
  variables.
\newblock \emph{J. Mach. Learn. Res.}, 6:1453--1484.

\bibitem[{Vaswani et~al.(2017)Vaswani, Shazeer, Parmar, Uszkoreit, Jones,
  Gomez, Kaiser, and Polosukhin}]{vaswani17attention}
Ashish Vaswani, Noam Shazeer, Niki Parmar, Jakob Uszkoreit, Llion Jones,
  Aidan~N. Gomez, Lukasz Kaiser, and Illia Polosukhin. 2017.
\newblock Attention is all you need.
\newblock In \emph{Advances in Neural Information Processing Systems 30: Annual
  Conference on Neural Information Processing Systems}, pages 5998--6008.

\bibitem[{Welleck and Cho(2020)}]{Welleck2020MLEguidedPS}
S.~Welleck and Kyunghyun Cho. 2020.
\newblock Mle-guided parameter search for task loss minimization in neural
  sequence modeling.
\newblock \emph{ArXiv}, abs/2006.03158.

\bibitem[{White et~al.(1994)White, O'Connell, and O'Mara}]{White1994TheAM}
J.~White, T.~O'Connell, and Francis~E. O'Mara. 1994.
\newblock The arpa mt evaluation methodologies: Evolution, lessons, and future
  approaches.
\newblock In \emph{AMTA}.

\bibitem[{Wieting et~al.(2019)Wieting, Berg-Kirkpatrick, Gimpel, and
  Neubig}]{Wieting2019BeyondBT}
J.~Wieting, Taylor Berg-Kirkpatrick, Kevin Gimpel, and Graham Neubig. 2019.
\newblock Beyond bleu: Training neural machine translation with semantic
  similarity.
\newblock In \emph{ACL}.

\bibitem[{Wiseman and Rush(2016)}]{Wiseman2016SequencetoSequenceLA}
Sam Wiseman and Alexander~M. Rush. 2016.
\newblock Sequence-to-sequence learning as beam-search optimization.
\newblock In \emph{EMNLP}.

\bibitem[{Wu et~al.(2018)Wu, Tian, Qin, Lai, and Liu}]{Wu2018ASO}
Lijun Wu, Fei Tian, Tao Qin, J.~Lai, and T.~Liu. 2018.
\newblock A study of reinforcement learning for neural machine translation.
\newblock In \emph{EMNLP}.

\bibitem[{Wu et~al.(2016)Wu, Schuster, Chen, Le, Norouzi, Macherey, Krikun,
  Cao, Gao, Macherey, Klingner, Shah, Johnson, Liu, Kaiser, Gouws, Kato, Kudo,
  Kazawa, Stevens, Kurian, Patil, Wang, Young, Smith, Riesa, Rudnick, Vinyals,
  Corrado, Hughes, and Dean}]{Wu2016GooglesNM}
Y.~Wu, M.~Schuster, Z.~Chen, Quoc~V. Le, Mohammad Norouzi, Wolfgang Macherey,
  M.~Krikun, Yuan Cao, Q.~Gao, Klaus Macherey, J.~Klingner, Apurva Shah,
  M.~Johnson, X.~Liu, Lukasz Kaiser, Stephan Gouws, Y.~Kato, Taku Kudo,
  H.~Kazawa, K.~Stevens, George Kurian, Nishant Patil, W.~Wang, C.~Young,
  J.~Smith, Jason Riesa, Alex Rudnick, Oriol Vinyals, G.~Corrado, Macduff
  Hughes, and J.~Dean. 2016.
\newblock Google's neural machine translation system: Bridging the gap between
  human and machine translation.
\newblock \emph{ArXiv}, abs/1609.08144.

\bibitem[{Zhang et~al.(2020)Zhang, Kishore, Wu, Weinberger, and
  Artzi}]{Zhang2020BERTScoreET}
Tianyi Zhang, V.~Kishore, Felix Wu, Kilian~Q. Weinberger, and Yoav Artzi. 2020.
\newblock Bertscore: Evaluating text generation with bert.
\newblock \emph{ArXiv}, abs/1904.09675.

\bibitem[{Zhu et~al.(2020)Zhu, Xia, Wu, He, Qin, Zhou, Li, and
  Liu}]{Zhu2020IncorporatingBI}
Jinhua Zhu, Yingce Xia, Lijun Wu, Di~He, Tao Qin, W.~Zhou, H.~Li, and T.~Liu.
  2020.
\newblock Incorporating bert into neural machine translation.
\newblock \emph{ArXiv}, abs/2002.06823.

\end{thebibliography}
\bibliographystyle{acl_natbib}

\appendix

\begin{figure*}[h]
\centering
  \includegraphics[width=\linewidth]{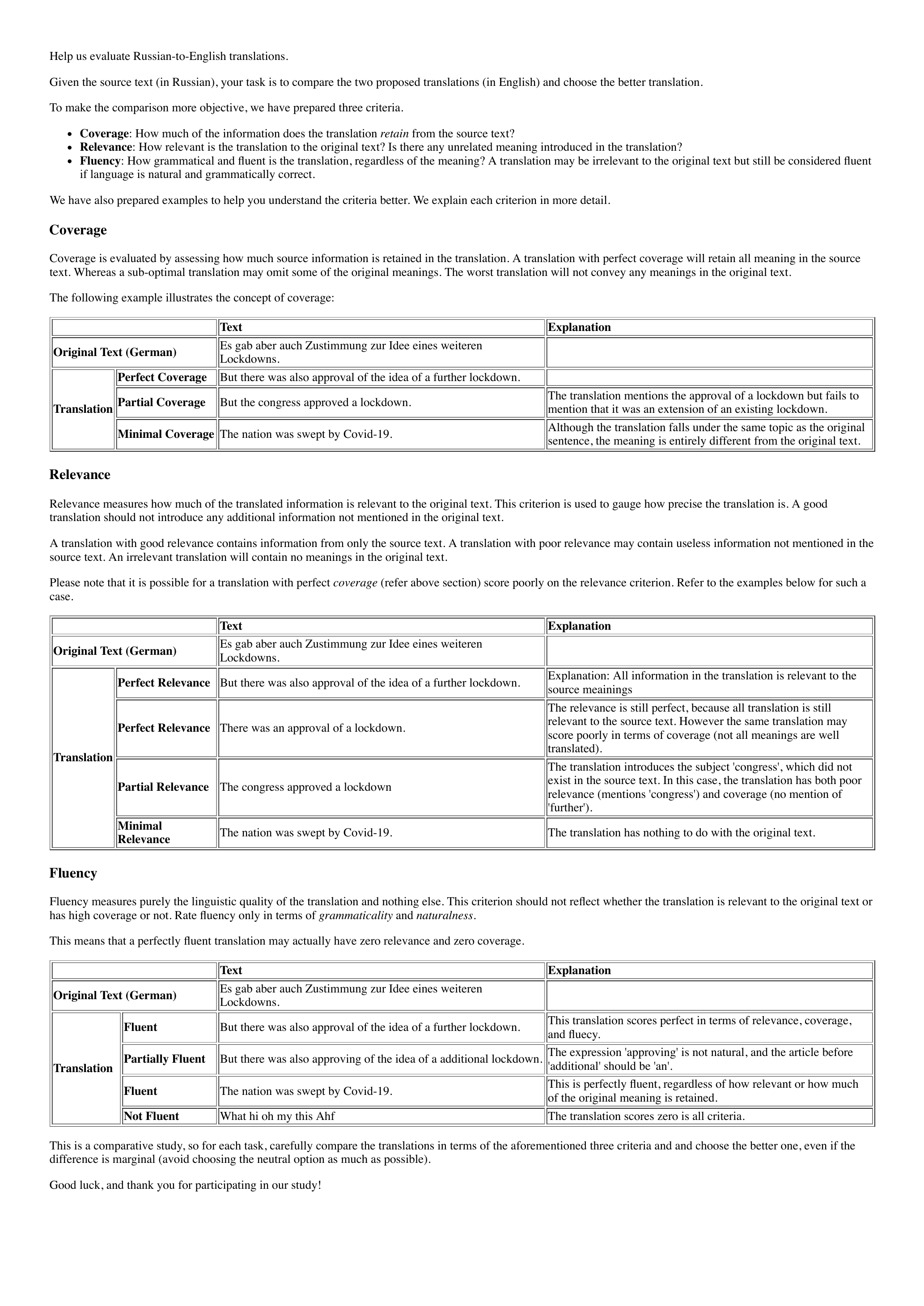}
\caption{Instruction page for pairwise human evaluation}
\label{fig:survey_1}
\end{figure*}

\begin{figure*}[h]
\centering
  \includegraphics[trim=0.5cm 5cm 2cm 0.5cm, width=\linewidth]{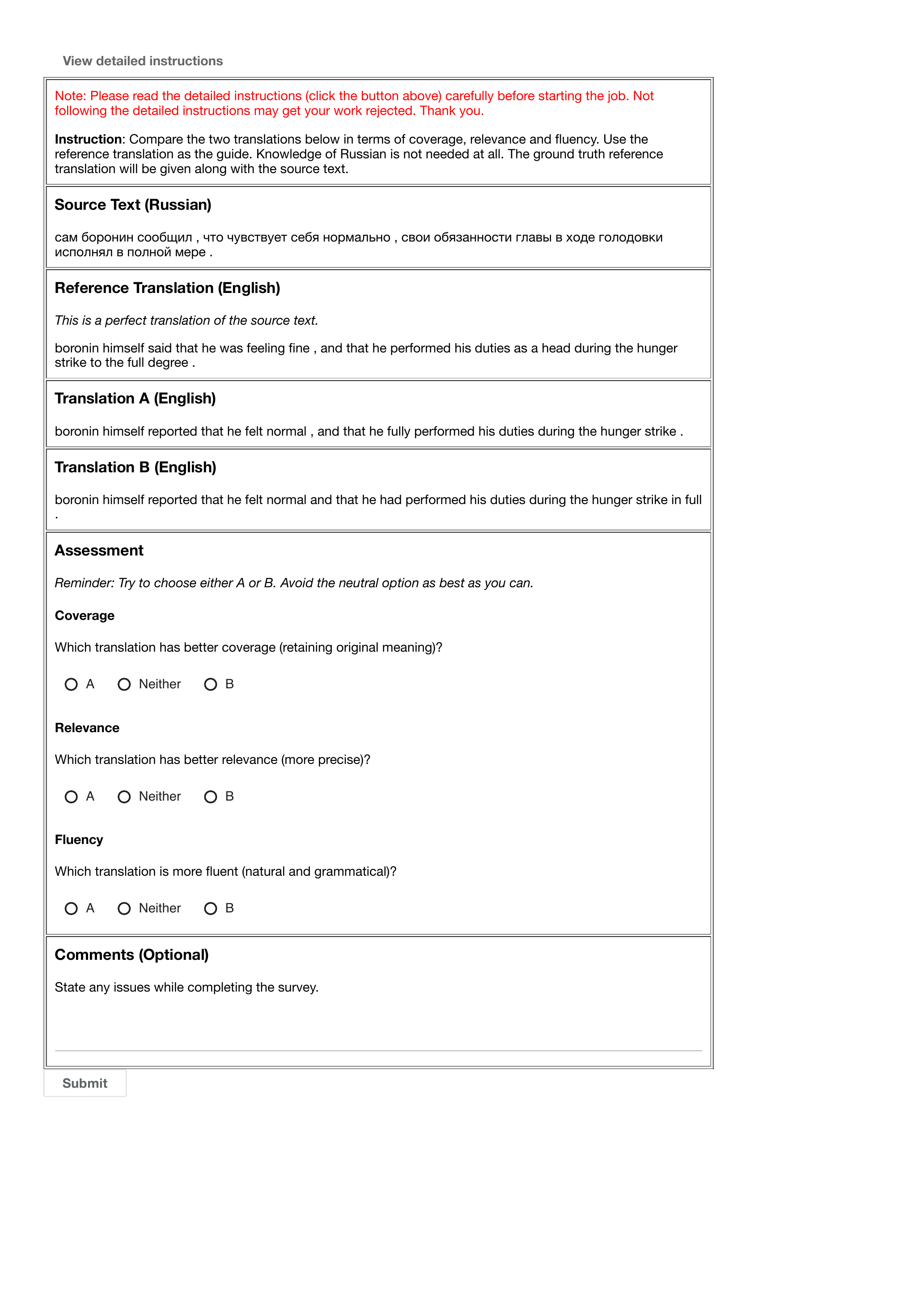}
\caption{Work sheet for pairwise human evaluation. This figure displays an example page for Russian-English language pair.}
\label{fig:survey_2}
\end{figure*}

\section{Pairwise Human Evaluation}

Fig.~\ref{fig:survey_1} and Fig.~\ref{fig:survey_2} show the questionnaire we give to annotators on Amazon Mechanical Turk.  In Fig.~\ref{fig:survey_1}, we display the instruction of scoring, and the detailed explanation for three quality criteria: coverage, adequacy (relevance), and fluency. All annotators will first read the instruction before the actual assessment.  Fig.~\ref{fig:survey_1} shows the assessment worksheet, which lists the source text, reference translation, and two translations from a random model pair. We do not require the annotators to understand the source language as the high-quality worker pool will be limited with the language filter. Therefore, the judgment is expected to be done by comparing with the reference translation.

\end{document}